\documentclass[letterpaper, 10 pt, conference]{ieeeconf}  

\IEEEoverridecommandlockouts                              

\title{\LARGE \bf
Tacmap: Bridging the Tactile Sim-to-Real Gap via Geometry-Consistent Penetration Depth Map
}

\author{Lei Su$^{*,1}$, Zhijie Peng$^{*,1,2}$, Renyuan Ren$^{1}$, Shengping Mao$^{1}$, Juan Du$^{3}$, Kaifeng Zhang$^{1}$, Xuezhou Zhu$^{1}$ \\
$^1$Sharpa \qquad \qquad $^2$HKUST \qquad \qquad $^3$NVIDIA \\
$^*$Equal Contribution
}

\usepackage{amsmath}
\usepackage{bm}
\usepackage{amssymb}
\usepackage{graphicx}
\usepackage{caption}
\usepackage{subfigure}
\usepackage{xcolor}

\begin{document}

\twocolumn[{%
\renewcommand\twocolumn[1][]{#1}%

\maketitle
\begin{center}
    \includegraphics[width=1.0\textwidth]{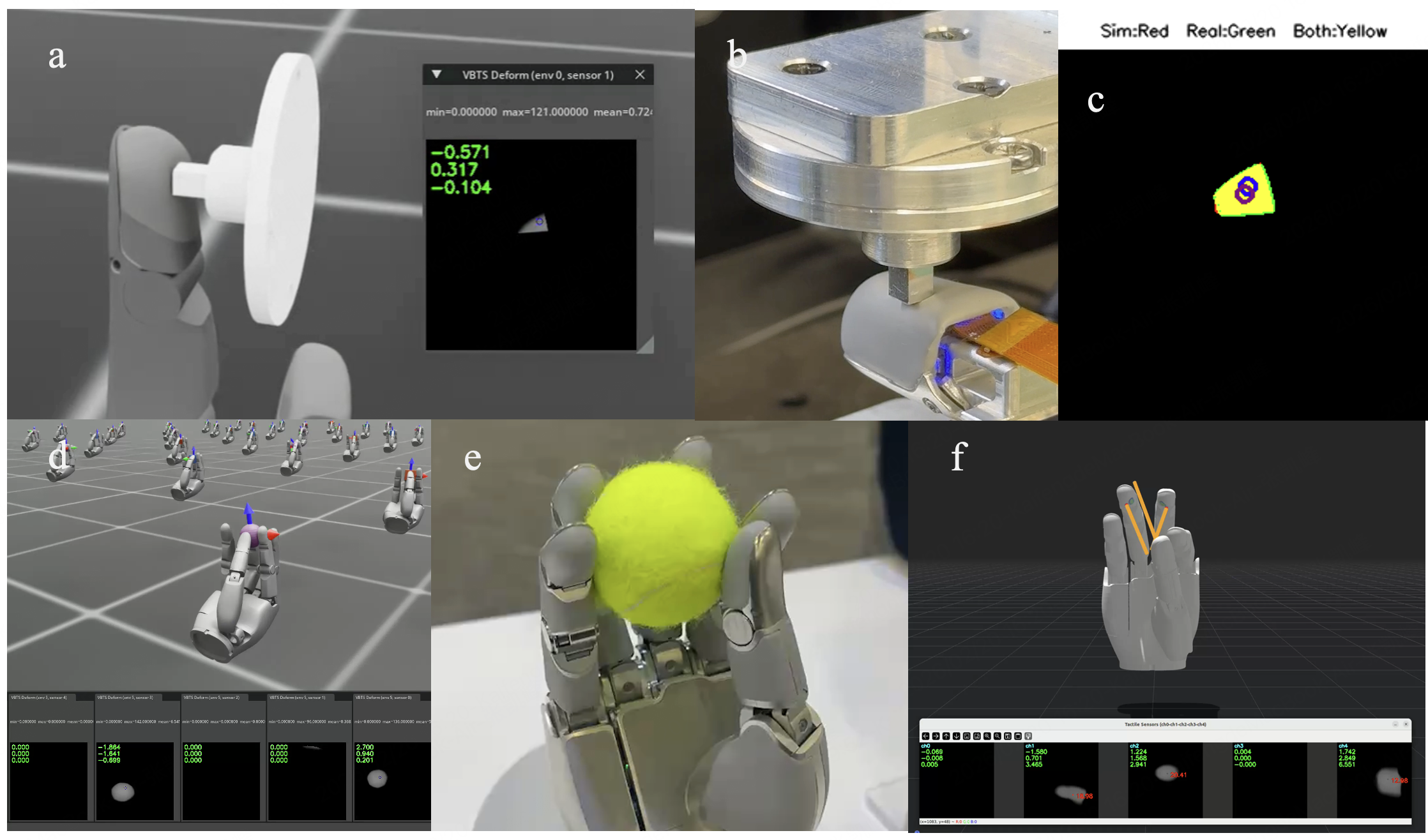}

    \captionof{figure}{(a-b) Standard setups for sim-to-real gap evaluation in (a) simulation and (b) the real world. (c) Comparison between simulation and real-world deform maps. (d-e) Deployment of an RL-based ball rotation policy from (d) simulation to (e) the real world. (f) Replaying real-world interaction data within the simulation environment for verification.}
    \label{fig:teaser}
\end{center}
}]

\begin{abstract}

Vision-Based Tactile Sensors (VBTS) are essential for achieving dexterous robotic manipulation, yet the tactile sim-to-real gap remains a fundamental bottleneck. Current tactile simulations suffer from a persistent dilemma: simplified geometric projections lack physical authenticity, while high-fidelity Finite Element Methods (FEM) are too computationally prohibitive for large-scale reinforcement learning. In this work, we present Tacmap, a high-fidelity, computationally efficient tactile simulation framework anchored in volumetric penetration depth. Our key insight is to bridge the tactile sim-to-real gap by unifying both domains through a shared deform map representation. Specifically, we compute 3D intersection volumes as depth maps in simulation, while in the real world, we employ an automated data-collection rig to learn a robust mapping from raw tactile images to ground-truth depth maps. By aligning simulation and real-world in this unified geometric space, Tacmap minimizes domain shift while maintaining physical consistency. Quantitative evaluations across diverse contact scenarios demonstrate that Tacmap's deform maps closely mirror real-world measurements. Moreover, we validate the utility of Tacmap through an in-hand rotation task, where a policy trained exclusively in simulation achieves zero-shot transfer to a physical robot.
\end{abstract}

\section{INTRODUCTION}

Vision-based tactile sensors (VBTS), such as GelSight \cite{yuan2017gelsight} and DIGIT \cite{lambeta2020digit}, have emerged as a pivotal technology in robotic manipulation. By leveraging high-resolution cameras to capture the deformation of an elastomer membrane, these sensors provide rich, multi-modal information including geometry \cite{li2014localization}, contact force \cite{yuan2017measurement}, and shear strain \cite{dong2018force}. Unlike traditional taxel-based sensors \cite{dahiya2009tactile}, vision-based tactile sensing offers the spatial resolution necessary for perceiving fine textures \cite{luo2018vitac} and complex geometric features, which is essential for achieving human-like dexterity in tasks such as edge following \cite{lepora2019pixels}, slip detection \cite{james2018slip}, and precise in-hand manipulation \cite{calandra2018more}.

Despite their potential, the widespread application of these sensors is hindered by the prohibitive cost of real-world data collection. Training robust robotic policies often requires millions of interaction samples, making high-fidelity simulation an indispensable tool \cite{zhao2020lvis}. A tactile simulation toolkit that simultaneously achieves computational efficiency and physical realism is crucial for the robotics community. Such a tool not only enables large-scale reinforcement learning in a sim-to-real pipeline \cite{ding2021sim} but also allows researchers to iterate on sensor designs and manipulation methods without the risk of damaging hardware.

Current tactile simulation methodologies generally fall into three categories, each with significant trade-offs. Empirical methods like Taxim \cite{lin2022taxim} rely on supervising models with real-world datasets. However, their dependency on specific data distributions often leads to poor generalization toward novel object geometries. Analytical methods such as TACTO \cite{wang2022tacto} utilize simple depth-buffer rendering, which fails to capture the physical deformation and volume-preserving characteristics of the elastomer, resulting in a significant sim-to-real gap. While physics-based simulators \cite{zhong2024tacsl, TacEx} offer higher fidelity, their integration into RL pipelines is hindered by extreme computational costs and the difficulty of matching complex material parameters to real-world sensors. Furthermore, most existing simulation toolkits are implicitly designed for flat-surface sensors. When applied to curved fingertips, which are increasingly common in anthropomorphic hands, these methods struggle with projection distortions and the non-planar nature of elastomer deformation, limiting their versatility across different hardware form factors.

To bridge this gap, we present Tacmap, a simulation framework that achieves high-fidelity tactile rendering by aligning simulation and reality within a unified geometric deformation space. Crucially, Tacmap is designed to be geometry-agnostic, supporting both flat and curved sensor surfaces by performing calculations in a generalized normal-projection space. Our key insight is that while raw tactile images are sensor-specific and optically complex, the underlying deformation map serves as a universal proxy for contact physics.

Specifically, in simulation, we introduce an efficient geometric calculation method that computes the 3D penetration depth between the rigid object and the sensor's elastomer. This depth field is used to synthesize the Tacmap, capturing the essential deformation features without the overhead of FEM. In the real world, we develop an automated data-collection rig capable of measuring ground-truth deformation during physical interaction. Using this dataset, we train a translation model to map raw tactile images to their corresponding deform maps.

By standardizing both domains into the Tacmap representation, we eliminate the need to simulate complex optical effects, allowing for a seamless sim-to-real transition grounded in consistent geometric principles.

The contributions of this paper are:

\begin{itemize}
    \item We propose a novel tactile simulation approach that utilizes penetration depth as a unified geometric representation to bridge the sim-to-real gap.
    \item We introduce a generalized rendering pipeline that naturally supports curved tactile fingertips. By computing deformations in the local surface normal space, Tacmap eliminates the projection artifacts common in existing flat-surface simulators.
    \item Our Tacmap achieves a balance between the rendering speed of analytical shaders and the physical accuracy of FEM-based methods, supporting massive parallel training in modern physics engines.
    \item Experiments demonstrate that policies trained via Tacmap can be directly transferred to a physical robot, proving the Tacmap's effectiveness for complex, contact-rich manipulation.
\end{itemize}

\section{RELATED WORK}

The development of vision-based tactile sensors (VBTS), such as GelSight \cite{yuan2017gelsight} and DIGIT \cite{lambeta2020digit}, has significantly enhanced robot manipulation capabilities. To bridge the tactile sim-to-real gap, researchers have proposed various simulation frameworks categorized into three main streams: empirical \cite{lin2022taxim, hoffman2018gan}, analytical \cite{wang2022tacto, johnson1987contact}, and physics-based methods \cite{ma2018gelsight, TacEx}.

Early methods primarily relied on image-to-image translation or empirical mapping. Taxim \cite{lin2022taxim} utilized an example-based approach, combining linear elastic models with real-world calibration to generate highly realistic tactile images. However, these methods often suffer from limited generalization to novel object geometries. Conversely, TACTO \cite{wang2022tacto} provided a scalable solution by rendering tactile imprints using standard rasterization engines. However, it simplifies the elastomer deformation as a basic depth mapping, sacrificing physical fidelity.

High-performance simulators have recently shifted towards GPU acceleration to meet the data demands of reinforcement learning (RL). TacSL \cite{zhong2024tacsl} leverages NVIDIA's Isaac Gym to achieve massive parallelism, utilizing a tensorized depth-to-RGB mapping that offers over 200$\times$ speedup compared to CPU-based methods. While efficient, it typically models the elastomer as a simplified spring-damper system, which may lack local detail during complex interactions.

Physics-fidelity oriented works like TacEx \cite{TacEx} and Taccel \cite{liu2025taccel} have integrated advanced soft-body dynamics. TacEx incorporates GIPC for intersection-free contact modeling, while Taccel combines Affine Body Dynamics (ABD) with Incremental Potential Contact (IPC) to achieve unprecedented simulation speeds for multi-contact scenarios.

\begin{figure*}[t]
    \centering
    \includegraphics[width=0.8\textwidth]{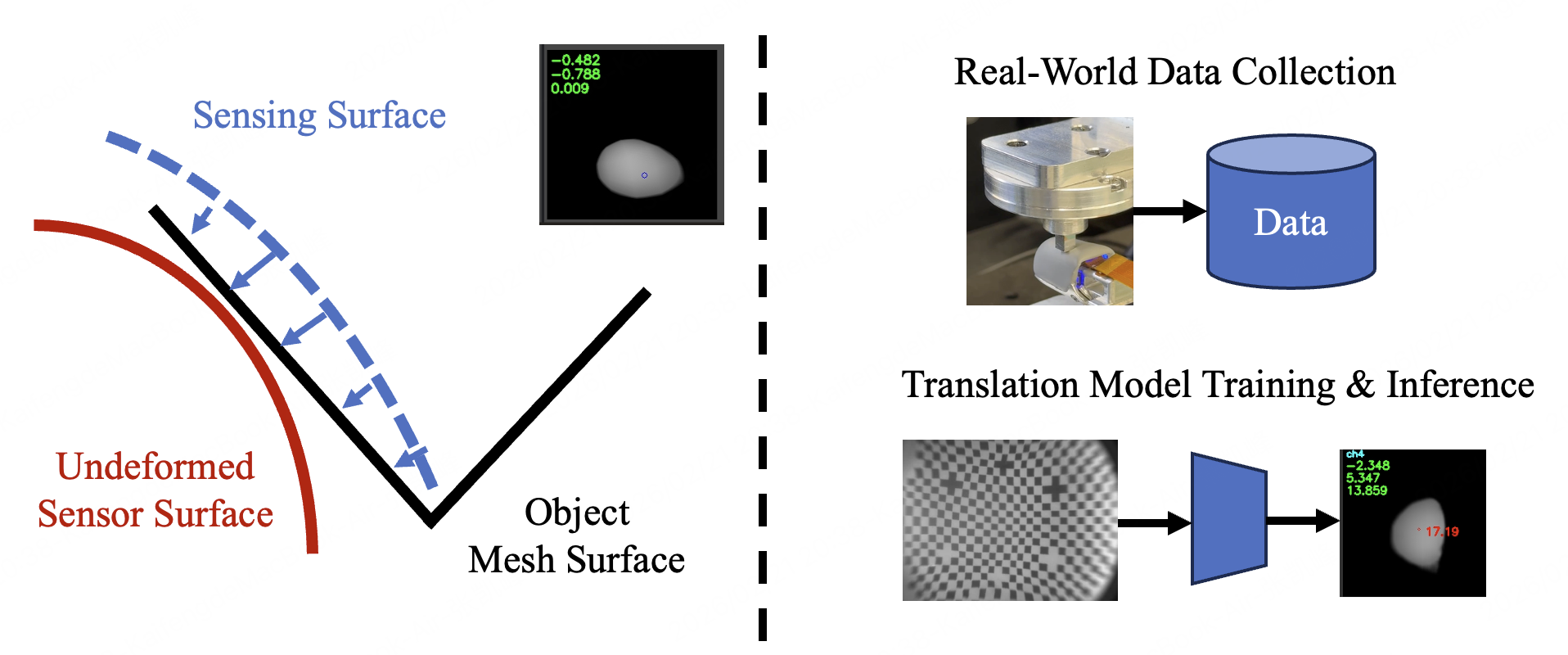} \\
    (a) Deform Map Generation in Simulation \qquad \qquad \quad (b) Deform Map Generation in Real-World
    \caption{Overview of the Tacmap Framework. (a) Diagram of deform map generation in simulation and (b) diagram of deform map generation in real-world.}
    \label{fig:method}
\end{figure*}

Despite these advancements, a critical gap remains in the geometric adaptability and computational-physical balance of current simulators. Our Tacmap bridges these gaps by introducing a unified geometric deformation space. It not only naturally accommodates diverse sensor geometries through normal-aligned depth calculation but also achieves a superior balance, providing higher physical consistency than simple shaders while maintaining the real-time rendering speeds essential for massive parallel training.

\section{METHOD}

\subsection{Overview}

In this work, we provide a comprehensive and synchronized tactile representation between simulation and real-world. To facilitate complex manipulation capabilities, we provide three primary streams of tactile information: net force $\mathbf{F}$, contact position $\mathbf{P}$, and deform map $\mathbf{M}$. Crucially, the deform map, i.e., Tacmap, effectively bridges the tactile sim-to-real gap, enabling high-fidelity alignment between simulation and real-world.

The acquisition of these signals is tailored to the characteristics of each domain to ensure geometry consistency.

\textbf{Net Force ($\mathbf{F}$)} In the simulation, the net force is directly retrieved from the physics engine's built-in contact sensors at the fingertip. In the real world, we collect a dataset of paired raw tactile images and ground-truth readings from an external high-precision force sensor. A ResNet-based regression network is then trained to map the raw tactile images directly to the estimated net force, enabling force perception using the vision-based tactile sensors. 

\textbf{Contact Position ($\mathbf{P}$)} In the simulation, $\mathbf{P}$ corresponds to the precise contact position reported by the contact sensor. In the real-world, we derive $\mathbf{P}$ by calculating the geometric centroid of the effective contact area within the predicted deform map $\mathbf{M}$.

\textbf{Deform Map ($\mathbf{M}$)} As the primary contribution in this work, i.e., Tacmap, the deform map represents the dense, pixel-wise penetration depth between the fingertip and the object. While $\mathbf{F}$ and $\mathbf{P}$ provide global interaction states, $\mathbf{M}$ captures high-resolution local geometry essential for fine-grained manipulation. Figure \ref{fig:method} depicts how Tacmap synthesizes deformation maps across the sim-to-real spectrum (detailed in Sec. \ref{method:real_world} - Sec. \ref{method:simulation}). Briefly, simulated maps are generated by calculating the 3D penetration depth during sensor-object interaction. For the real-world domain, we leverage an automated data-collection rig to train a translation model that transforms raw tactile frames into unified deformation maps, ensuring geometric consistency between both environments.

\subsection{Deform Map Generation in Simulation}\label{method:simulation}
To bypass the prohibitive computational cost of Finite Element Methods (FEM), we introduce a high-fidelity geometric rendering pipeline that characterizes the physical interaction between the sensor elastomer and rigid objects.

\textbf{Geometric Modeling} We define the sensor's elastomer geometry using two critical boundaries: the undeformed sensor surface $S_{u}$, representing the physical resting state of the fingertip, and a virtual sensing surface $S_{s}$. Crucially, $S_{s}$ is positioned at a fixed offset exterior to the sensor, acting as a virtual boundary that encapsulates the interaction zone. When a rigid object $\mathcal{O}$ approaches and contacts the sensor, it traverses the gap between these two surfaces, creating a multi-layered intersection volume.

\begin{figure*}[t]
    \centering
    \includegraphics[width=0.8\textwidth]{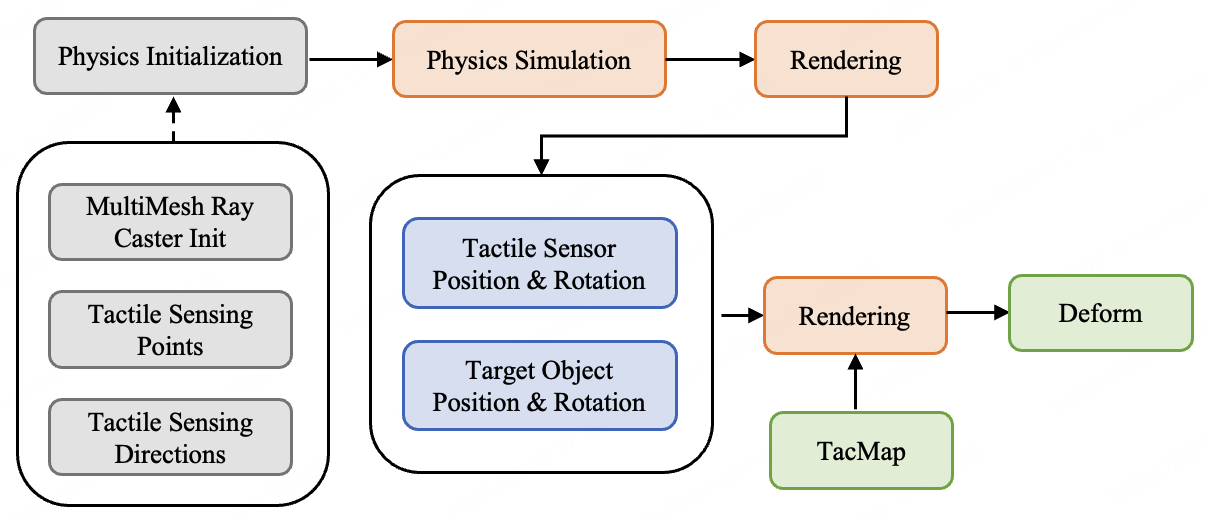}
    \caption{The implementation framework of our Tacmap in Isaac Lab and MuJoCo.}
    \label{fig:implementation}
\end{figure*}

\textbf{Deform Generation} We discretize the sensing surface $S_s$ into an $H \times W$ grid. For each grid point $(u, v)$, a projection ray $\mathbf{r}_{u,v}$ is cast from the sensing surface $S_s$ toward the sensor's interior, along the surface normal. The deformation value $d(u, v)$ is calculated based on the object's occupancy within the interval between $S_s$ and $S_u$. Formally:

$$d(u, v) = \max(0, z_{s} - \max(z_{u}, z_{o}))$$

where $z_s$ is the origin of the ray on the sensing surface, $z_u$ is the coordinate of the physical undeformed sensor surface, and $z_o$ is the coordinate of the first intersection point between the ray $\mathbf{r}_{u,v}$ and the object mesh $\mathcal{O}$. Note that $S_s$ and $S_u$ are almost the same one.

The primary advantage of the Tacmap representation is its domain-invariant nature. By abstracting tactile feedback into a purely geometric deform map $\mathbf{M}$, we decouple the underlying contact physics from sensor-specific optical artifacts, such as internal reflections and light scattering. This unification ensures that simulated tactile signals are structurally congruent with physical measurements. Consequently, Tacmap mitigates the sim-to-real gap by aligning both domains in a shared geometric space, providing a computationally efficient alternative for large-scale reinforcement learning without sacrificing physical plausibility.

\subsection{Deform Map Generation in Real-World} \label{method:real_world}
To enable the robot to perceive the same deformation representation in the physical world as in simulation, we develop a translation pipeline that infers deform maps directly from raw tactile frames. This process involves utilizing an automated workstation to generate a paired dataset of tactile images and their corresponding geometric ground-truth, which is then used to train a robust mapping function.

\textbf{Automated Data Collection} We developed an automated hardware-in-the-loop data collection system to capture the elastomer's response under controlled conditions. We employ a high-precision 3-axis motion stage, i.e., Figure \ref{fig:method}, to orchestrate interactions between the sensor and a diverse set of geometric indenters. For each compression, the indenter's precise 3D pose, $\mathbf{T}_{\text{tool}}$, is recorded via the stage's encoders relative to the sensor’s reference frame. By using the similar techniques as simulation, we ensure that the real-world deform maps are calculated using the exact same geometric projection logic as the simulated maps in Sec. 2.2, forming a synchronized dataset $\mathcal{D} = \{(\mathbf{I}_{\text{raw}}^{(i)}, \mathbf{M}_{gt}^{(i)})\}_{i=1}^N$.

\textbf{Image-to-Deform Translation} With the collected dataset, we train a translation network $\Phi$ to generate the deform map from a single tactile image: $\mathbf{\hat{M}} = \Phi(\mathbf{I}_{\text{raw}})$. 

We adopt a ResNet-based encoder-decoder architecture, optimized to invert the complex, non-linear optical phenomena, such as internal reflection and light scattering, back into the underlying geometric representation. The network is trained by minimizing the pixel-wise mean squared error (MSE) between the predicted $\mathbf{\hat{M}}$ and the kinematically-derived $\mathbf{M}_{gt}$.

\textbf{Rationale for Sim-to-Real Alignment} The efficacy of Tacmap in minimizing the sim-to-real gap stems from two factors. First, by supervising the real-world network with geometric depth rather than raw images, we decouple the sensor's optical characteristics from the contact physics. Second, because both the simulated $\mathbf{M}_{\text{sim}}$ and the real-world $\mathbf{\hat{M}}$ are anchored to the same physical reference (the penetration depth), they reside in a shared, domain-invariant space. This alignment allows policies trained in simulation to interpret real-world tactile signals without the characteristic domain shift associated with raw pixel intensities.

\section{IMPLEMENTATION}

The simulation toolkit Tacmap is integrated into both Isaac Lab and MuJoCo to ensure broad compatibility and scalability. In this section, we introduce the implementation details of our Tacmap as follows. Figure \ref{fig:implementation} shows the implementation framework of Tacmap. 

\begin{figure*}[t]
    \centering
        \begin{minipage}{0.4\textwidth}
        \centering
        \includegraphics[width=\textwidth]{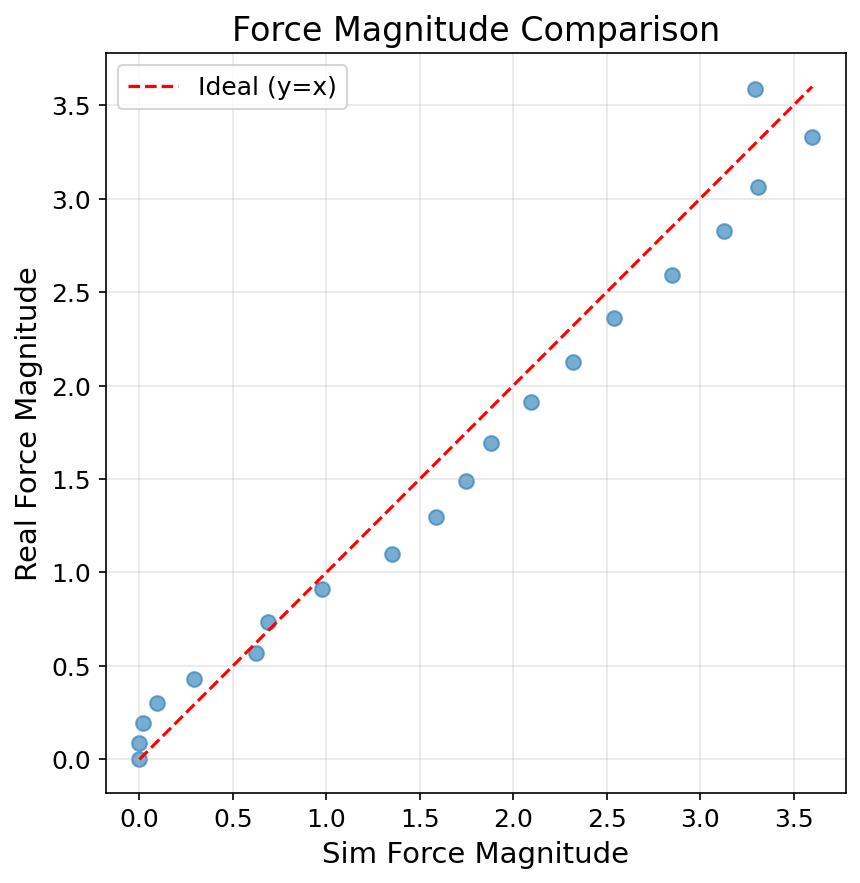}
    \end{minipage}
    \begin{minipage}{0.4\textwidth}
        \centering
        \includegraphics[width=\textwidth]{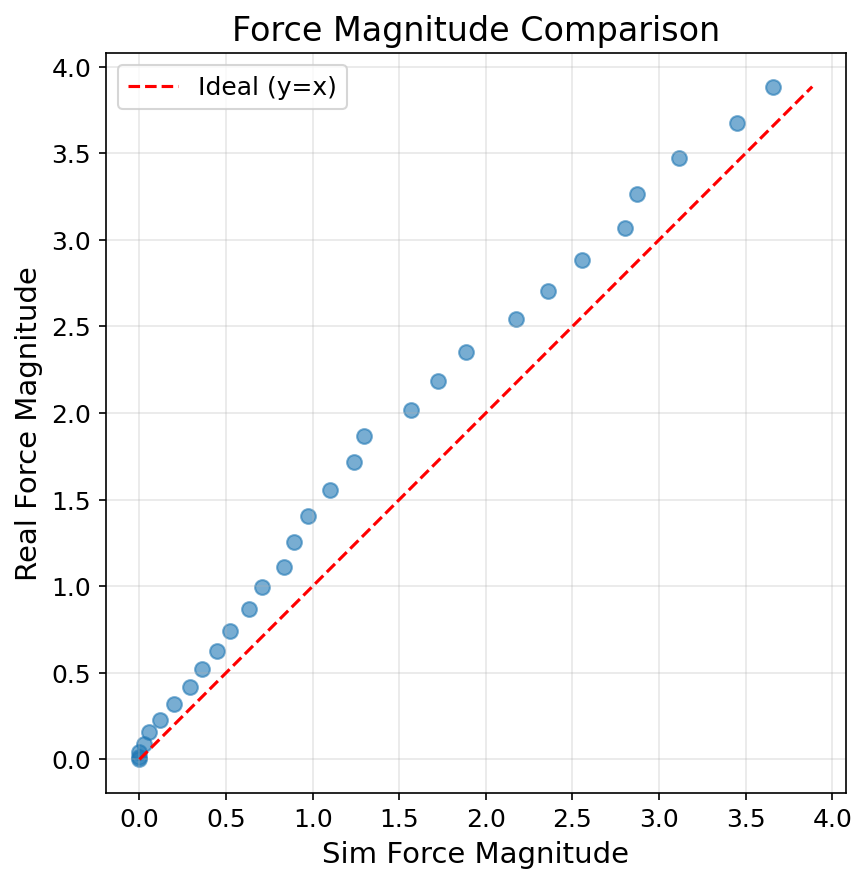}
    \end{minipage}
    \caption{Net force comparison between simulation and real-world under the same relative contact positions between object and fingertip (left: cylinder, right: square).}
            \label{fig:net_force}
\end{figure*}

During the system initialization phase, we leverage a Multi-Mesh Ray Caster that pre-defines a dense grid of Tactile Sensing Points and their corresponding Sensing Directions on the undeformed sensor surface. This design effectively decouples the geometric resolution of tactile sensing from the underlying physics collision meshes, enabling high-fidelity tactile feedback without compromising the numerical stability of the physics solver.

At runtime, the toolkit maintains precise synchronization between the poses of the Tactile Sensor and the Target Object. These states are fed into the rendering module, where GPU-accelerated ray-casting calculates the geometric interpenetration in parallel. Specifically, in Isaac Lab, we leverage the high-performance Raycaster API to support massive parallel rendering across thousands of simultaneous environments. In MuJoCo, we implement a dedicated toolkit utilizing the mj$\_$ray function to efficiently query penetration depths.

Finally, the raw intersection data is processed through a vectorized pipeline to synthesize the deform map. By offloading the entire rendering process to the GPU, Tacmap achieves a high frequency rendering while maintaining high structural alignment with real-world physical measurements. This implementation provides the computational efficiency necessary for large scale reinforcement learning while ensuring the geometric fidelity required for sim-to-real transfer in contact rich manipulation tasks.

\section{EXPERIMENT}

In this section, we present a comprehensive evaluation of the Tacmap toolkit. All physical experiments are conducted using the SharpaWave dexterous hand, equipped with its vision-based tactile sensors (i.e., DTC). Our experiment design aims to address the following three research questions:

\textbf{Sim-to-Real Fidelity} To what extent does the simulated Tacmap align with real-world tactile measurements across diverse contact scenarios?

\textbf{Computational Efficiency} Can the Tacmap generation pipeline maintain the high rendering throughput required for large-scale parallel reinforcement learning in environments like Isaac Lab?

\textbf{Task Effectiveness} Does the Tacmap provide sufficient geometric information to support the zero-shot sim-to-real transfer of complex, contact-rich manipulation policies, such as in-hand rotation?

\subsection{Experimental Setup}

\textbf{Hardware Platform} To validate the efficacy of Tacmap in complex manipulation tasks, we employ the SharpaWave dexterous hand as our primary hardware platform. The hand is equipped with high-speed actuation and vision-based tactile fingertips, which provide the raw sensory input for real-world deployment.

\textbf{Simulation Environment} We integrate the Tacmap toolkit into Isaac Lab and MuJoCo to create a high-fidelity tactile simulation environment. Rather than mere kinematic replication, we focus on the systematic alignment of the tactile sensor's physical properties. Specifically, we instantiate the dual-surface geometric model (as described in Sec. \ref{method:simulation}) within these physics engines, enabling real-time generation of deform maps during dynamic interactions.

\textbf{Object Dataset} To evaluate the generalization and robustness of Tacmap, we utilize a diverse library of objects. This variety allows us to benchmark the consistency of deform maps across different scales and surface curvatures.

\subsection{Empirical Study}

\begin{figure*}[t]
    \centering
    \includegraphics[width=\textwidth]{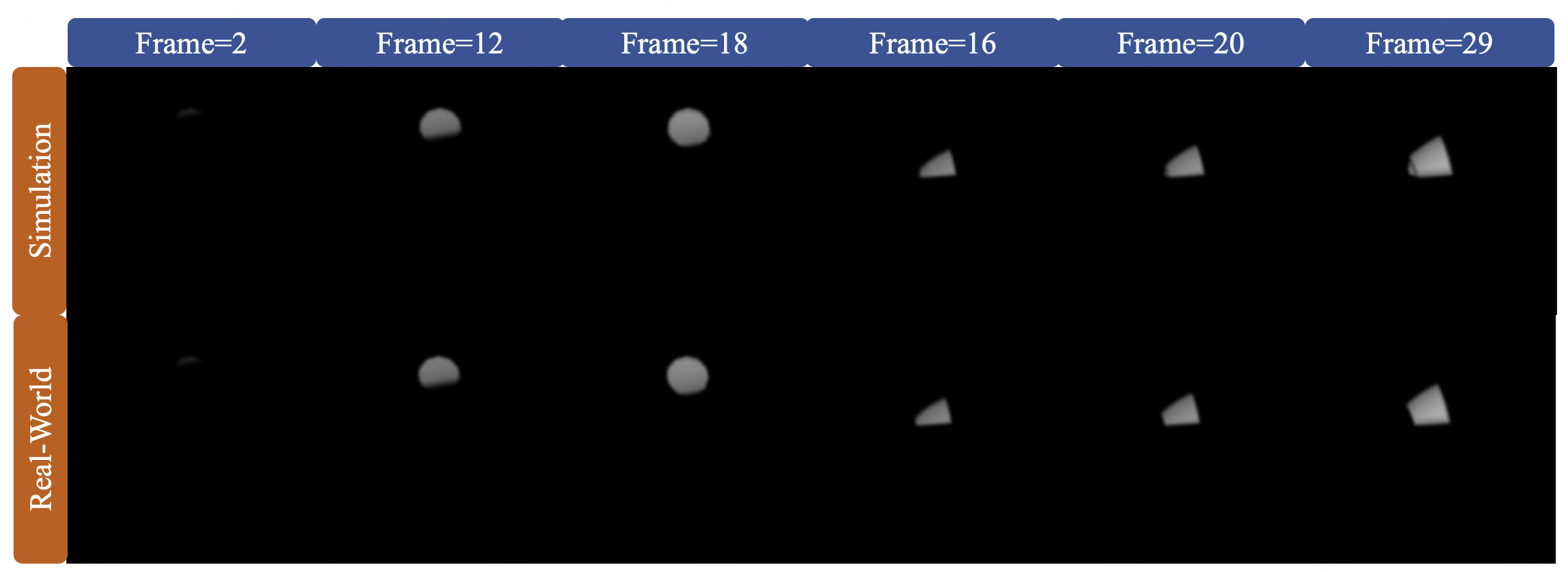}
    \caption{Visualization of deform map across simulation and real world under the same contact position with cylinder and square.}
            \label{fig:visualization}
\end{figure*}

\begin{table*}[h]
\caption{Quantitative evaluation of Tacmap between simulation and real-world. Note that all the values are the median value across diverse contact scenarios.}
\label{table:empirical_study}
\begin{center}
\begin{tabular}{|c||c||c||c||c|}
\hline
Object & Contact Position Error & Deform Depth Error &  Net Force $L_2$ Error & Deform IoU\\
\hline
Square & $0.66mm$ & $18.53\%$ & $0.28N$ & $88.21\%$ \\
\hline
Cylinder & $0.96mm$ & $14.71\%$ & $0.61N$ & $85.67\%$\\
\hline
\end{tabular}
\end{center}
\end{table*}

To evaluate the physical and geometric consistency of Tacmap, we conduct an empirical study comparing simulated contact against real-world measurements. The objective is to verify whether our unified deform map can simultaneously preserve global force dynamics and local contact geometry, both of which are critical for stable robotic manipulation. Figure \ref{fig:teaser} a-b shows the setups for sim-to-real gap evaluation.

\textbf{Force Alignment Analysis} We first investigate the alignment of the net force, which governs the high-level dynamics of robot object interaction. Using cylindrical and square indenters, we perform standardized compression tests. Figure \ref{fig:net_force} illustrates the correlation between the simulated force derived from the physics engine and the estimated force from the real-world sensor. The results indicate that the two domains are highly correlated, suggesting that our geometric penetration model serves as an effective proxy for the underlying contact mechanics. This alignment is crucial for tasks requiring precise force control or contact-rich transitions.

\textbf{Geometric Consistency} To assess the fine-grained fidelity of the deformation representation, we visualize the evolution of contact manifolds across six key timestamps during two compression process. As shown in Figure \ref{fig:visualization}, the side-by-side comparison reveals that the simulated deform maps and the real-world inferred deform maps exhibit remarkable structural similarity.

\textbf{Quantitative Assessment} Table \ref{table:empirical_study} provides a quantitative evaluation of the domain discrepancy. By adopting penetration depth as a unified representation, Tacmap successfully captures the essential geometric features of the contact manifold without the need for computationally expensive optical modeling. Quantitatively, Tacmap minimizes the sim-to-real gap, a level sufficient for zero-shot policy transfer.

These empirical results confirm that the proposed "Common Geometric Space" effectively bridges the domain gap. By ensuring consistency in both force and geometry, Tacmap provides a reliable foundation for downstream manipulation tasks, where the policy generalizes from simulated geometric abstractions to raw physical interactions.

\subsection{Computational Efficiency}

The integration of high-resolution tactile sensing into robot learning pipelines typically introduces two major computational bottlenecks: excessive GPU memory consumption and increased rendering latency. For large-scale parallel training, it is imperative that the tactile simulation remains computationally lightweight to avoid degrading the stepping frequency of the physics engine. In this section, we evaluate the scalability of Tacmap within the Isaac Lab framework as the number of parallel environments increases.

\begin{figure*}[t]
    \centering
        \begin{minipage}{0.4\textwidth}
        \centering
        \includegraphics[width=\textwidth]{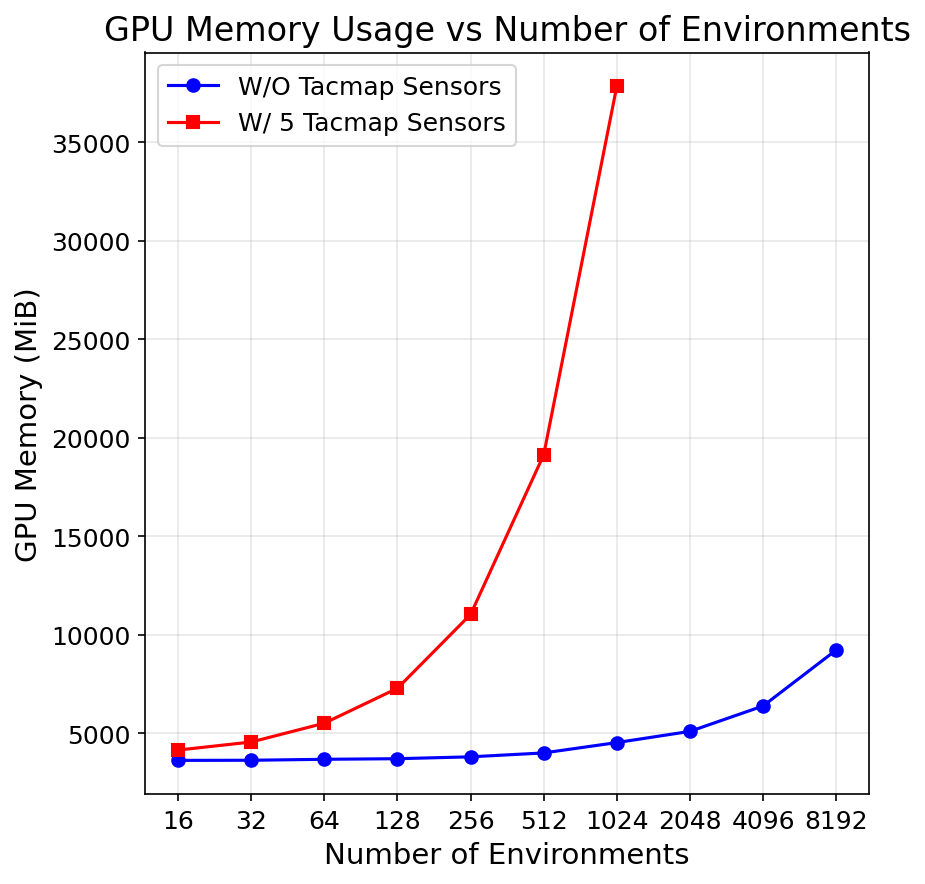}
    \end{minipage}
    \begin{minipage}{0.4\textwidth}
        \centering
        \includegraphics[width=\textwidth]{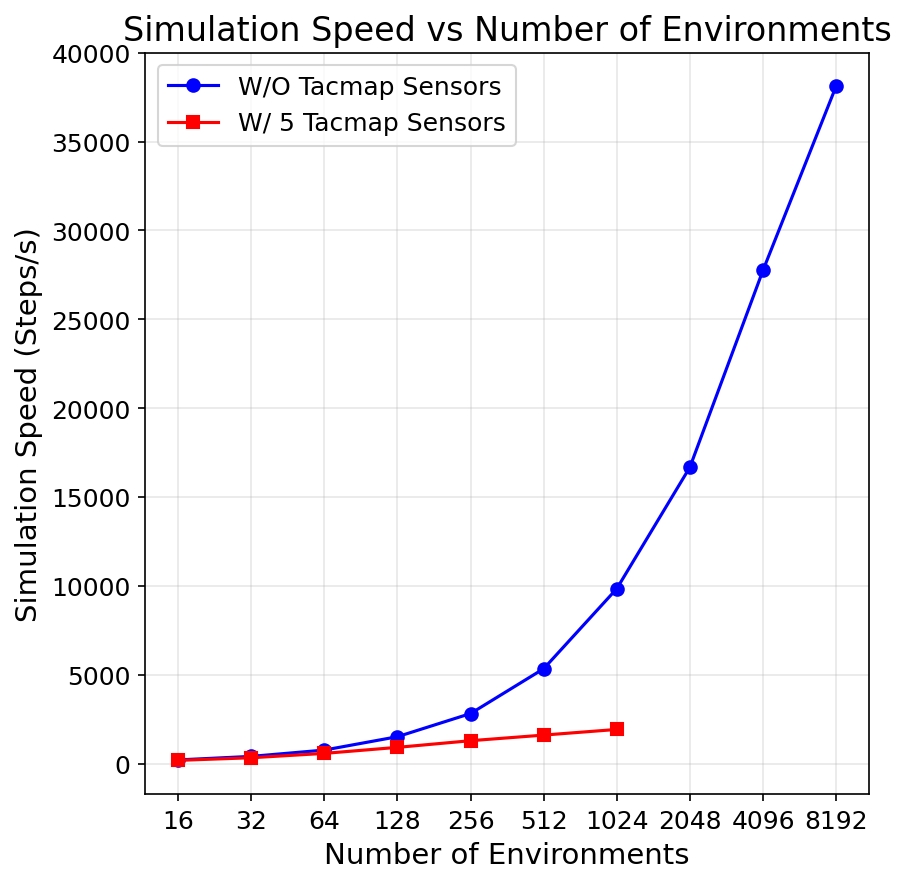}
    \end{minipage}
    \caption{Rendering efficiency of our Tacmap with GPU memory usage (left) and simulation rendering speed (right).}
            \label{fig:efficiency}
\end{figure*}

\textbf{GPU Memory Scalability} In tactile centric tasks, memory overhead often surges with the complexity of the sensor array. We monitored GPU allocation as environment scaled from 16 to 8192. Figure \ref{fig:efficiency} indicates that because Tacmap utilizes a geometric ray-casting approach rather than memory-intensive finite element method (FEM) calculations, the memory footprint exhibits only near-linear growth. This efficiency ensures that complex tactile-aware policies can be trained on a single consumer-grade GPU.

\textbf{Rendering Throughput} The second critical factor is the impact of rendering on the simulation's speed. We compared the total system throughput with and without Tacmap rendering enabled. Since our algorithm is fully executed on the GPU and integrated directly into the vectorized pipeline, the penetration depth calculation is highly synchronized with the physics steps. Figure \ref{fig:efficiency} shows that even with thousands of concurrent environments, Tacmap maintains a reasonable rendering frequency. Consequently, the inclusion of high-fidelity tactile information results in negligible degradation of the overall simulation speed, allowing RL agents to maintain peak data-sampling efficiency.

Overall, these results demonstrate that Tacmap successfully resolves the inherent conflict between high-fidelity tactile rendering and computational performance. By providing a scalable, low-latency feedback loop, the toolkit is well-suited for modern large-scale reinforcement learning frameworks, providing real-time tactile data without compromising computational resources.

\subsection{Sim-to-Real Application}

\textbf{Task Definition} To demonstrate the practical utility of Tacmap, we evaluate its performance on the task of in-hand object rotation, a complex manipulation challenge that requires continuous re-orientation while maintaining grasp stability. This task is particularly demanding as it requires the agent to perceive subtle contact transitions and modulate forces in real-time based on high-resolution tactile feedback.

\textbf{Policy Training} We train the control policy exclusively in simulation using Proximal Policy Optimization (PPO). The agent's observation space includes the real-time Tacmap stream, providing dense geometric information about the contact manifold. Leveraging the high-throughput rendering capability of our toolkit, the agent is exposed to millions of diverse contact interactions. This large-scale training allows the policy to internalize the geometric gradients of the deform maps, learning to anticipate and counteract potential slips through proactive finger coordination.

\textbf{Zero-Shot Deployment} The trained policy was deployed directly onto the physical SharpaWave dexterous hand without any real-world fine-tuning or domain adaptation. We tested the system using a spherical object to evaluate its ability to achieve smooth, continuous rotation. Experimental results show that the agent successfully interprets real-world tactile images—translated into deform maps via the translation model.

The successful execution of the in-hand rotation task provides definitive evidence of the physical fidelity and robustness of the Tacmap framework. By providing a physically grounded and domain-invariant bridge, Tacmap enables the transfer of sophisticated manipulation skills from simulation to real-world with high consistency.

\section{DISCUSSION}

In this work, we introduce Tacmap, a unified tactile representation framework designed to bridge the gap between simulation and the real world. By prioritizing geometric consistency and computational speed, Tacmap provides a scalable solution for the robotics community to develop tactile-aware policies that can transition seamlessly from virtual training to real-world deployment on dexterous hand. While our results demonstrate its effectiveness in contact-rich manipulation, several aspects regarding its implementation and future work merit further discussion.

\textbf{Advantages}
The primary advantage of Tacmap is the strict geometric alignment achieved between hardware and simulation. By standardizing tactile feedback into a "Common Geometric Space" of deform maps, we circumvent the traditional need for complex and sensor-specific optical modeling. This decoupling allows the simulation to remain agnostic to internal lighting or camera noise while providing the physical robot with a representation that is structurally identical to the synthetic data used during training. This alignment is the key driver behind the zero-shot transfer success observed in our dexterous manipulation tasks, as the policy learns to react to pure contact geometry rather than visual artifacts. 

\textbf{Limitations}
Despite its advantages, the current version of Tacmap has several limitations that provide clear avenues for future research. First, Tacmap focuses on normal penetration depth to generate deform maps. It does not explicitly model the tangential force distribution (shear strain) across the elastomer surface. Capturing these micro-slips and lateral forces is essential for even higher-level dexterity, such as predicting incipient slip during delicate picking. Future work will involve a co-design approach, where we iterate on both the hardware sensor and the Tacmap framework to incorporate a multi-dimensional force field representation.

Finally, although our implementation achieves computational efficiency, there remains significant room for optimization. As the complexity of the object meshes increases, the overhead of ray-casting would grow. We plan to explore more advanced acceleration structures, to further reduce the latency of penetration depth calculations in massive parallel environments.

\section{CONCLUSION}

In this paper, we present Tacmap, a high-fidelity and computationally efficient tactile simulation toolkit designed to bridge the sim-to-real gap for vision-based tactile sensors (VBTS). By introducing a unified deform map representation, we established a common geometric language that aligns simulation penetration depth with real-world physical measurements. Our Tacmap methodology ensures that tactile feedback is both physically consistent and scalable for massive parallel training. Empirical evaluations confirm that Tacmap achieves high structural similarity between simulation and real-world, while maintaining the rendering throughput required for large-scale reinforcement learning. Furthermore, the successful zero-shot transfer of a complex in-hand rotation policy to the physical dexterous hand demonstrates that our geometric representation captures the essential contact features necessary for high-level dexterity.

\bibliographystyle{IEEEtran}
\bibliography{root}

@inproceedings{lin2022taxim,
  title={Taxim: An example-based simulation framework for tactile finger photo-realism},
  author={Lin, Shaoxiong and Gu, Alex and Alspach, Alexander and Isola, Phillip},
  booktitle={2022 IEEE International Conference on Robotics and Automation (ICRA)},
  pages={10156--10162},
  year={2022},
  organization={IEEE}
}

@inproceedings{hoffman2018gan,
  title={CyCADA: Cycle-consistent adversarial domain adaptation},
  author={Hoffman, Judy and Tzeng, Eric Christie and Park, Taesung and Zhu, Jun-Yan and Isola, Phillip and Saenko, Kate and Efros, Alexei and Darrell, Trevor},
  booktitle={International Conference on Machine Learning (ICML)},
  pages={1989--1998},
  year={2018},
  organization={PMLR}
}

@inproceedings{wang2022tacto,
  title={TACTO: A Flexible, Open-source Simulator for High-resolution Vision-based Tactile Sensors},
  author={Wang, Sudharshan and Lambeta, Mike and Chou, Po-Wei and Calandra, Roberto},
  booktitle={2022 IEEE International Conference on Robotics and Automation (ICRA)},
  pages={5--11},
  year={2022},
  organization={IEEE}
}

@book{johnson1987contact,
  title={Contact mechanics},
  author={Johnson, Kenneth Langstreth},
  year={1987},
  publisher={Cambridge University Press}
}

@inproceedings{ma2018gelsight,
  title={Simulating gelsight tactile sensors for capturing geometry and force information},
  author={Ma, Jiachen and Donlon, Edward and Sanneman, Lindsay and Adelson, Edward H},
  booktitle={2018 IEEE/RSJ International Conference on Intelligent Robots and Systems (IROS)},
  pages={1031--1037},
  year={2018},
  organization={IEEE}
}

@article{zhong2024tacsl,
  title={TacSL: A Tensorized Tactile Simulation Library for Accelerating Robot Learning},
  author={Zhong, Siyuan and Hu, Hejia and Xu, Jing and Fang, Bin},
  journal={IEEE Robotics and Automation Letters},
  volume={9},
  number={4},
  pages={3156--3163},
  year={2024},
  publisher={IEEE}
}

@article{TacEx,
      title={TacEx: GelSight Tactile Simulation in Isaac Sim -- Combining Soft-Body and Visuotactile Simulators},
      author={Duc Huy Nguyen and Tim Schneider and Guillaume Duret and Alap Kshirsagar and Boris Belousov and Jan Peters},
      year={2024},
      eprint={2411.04776},
      archivePrefix={arXiv},
      primaryClass={cs.RO},
      url={https://arxiv.org/abs/2411.04776},
}

@inproceedings{liu2025taccel,
  title={Taccel: Accelerating High-Fidelity Tactile Simulation with Affine Body Dynamics and IPC},
  author={Liu, Minghua and Zhang, Yuzhe and Li, Ziyue and Wang, He Clarice and Yi, Li and Su, Hao},
  booktitle={2025 IEEE International Conference on Robotics and Automation (ICRA)},
  year={2025}
}

@article{yuan2017gelsight,
  title={GelSight: High-resolution robot tactile sensors for in-hand manipulation},
  author={Yuan, Wenzhen and Dong, Siyuan and Adelson, Edward H},
  journal={Sensors},
  volume={17},
  number={12},
  pages={2762},
  year={2017},
  publisher={MDPI}
}

@inproceedings{lambeta2020digit,
  title={DIGIT: A Finger-Sized High-Resolution Tactile Sensor for Dexterous Manipulation},
  author={Lambeta, Mike and Chou, Po-Wei and Tian, Stephen and Yang, Brian and Benjamin, Ivan and Dave, Abhinav and Piacenza, Claudio and Ma, Jiachen and Zhang, Siyuan and Fan, Linxi and Hausman, Karol and Righetti, Ludovic and Adelson, Edward H. and Calandra, Roberto},
  booktitle={2020 IEEE International Conference on Robotics and Automation (ICRA)},
  pages={5888--5895},
  year={2020},
  organization={IEEE}
}

@inproceedings{li2014localization,
  title={Localization and manipulation of small parts using GelSight tactile sensing},
  author={Li, Rui and Platt, Robert and Yuan, Wenzhen and Tenzer, Andreas and Adelson, Edward H},
  booktitle={2014 IEEE International Conference on Robotics and Automation (ICRA)},
  pages={3988--3993},
  year={2014}
}

@inproceedings{yuan2017measurement,
  title={Measurement of shear and normal forces on an active tactile sensor},
  author={Yuan, Wenzhen and Zhu, Chenzhuo and Owens, Andrew and Srinivasan, Mandayam A and Adelson, Edward H},
  booktitle={2017 IEEE International Conference on Robotics and Automation (ICRA)},
  pages={4446--4453},
  year={2017}
}

@inproceedings{dong2018force,
  title={Force estimation and slip detection/prediction for wrap-around gripper using GelSight tactile sensor},
  author={Dong, Siyuan and Yuan, Wenzhen and Adelson, Edward H},
  booktitle={2018 IEEE International Conference on Robotics and Automation (ICRA)},
  pages={1--8},
  year={2018}
}

@article{dahiya2009tactile,
  title={Tactile sensing—from humans to humanoids},
  author={Dahiya, Ravinder S and Metta, Giorgio and Valle, Maurizio and Sandini, Giulio},
  journal={IEEE Transactions on Robotics},
  volume={26},
  number={1},
  pages={1--20},
  year={2009}
}

@inproceedings{luo2018vitac,
  title={ViTac: Alive tactile sensing},
  author={Luo, Shan and Bimbo, Joao and Dahiya, Ravinder and Liu, Hongbin},
  booktitle={2018 IEEE International Conference on Robotics and Automation (ICRA)},
  pages={1--9},
  year={2018}
}

@article{lepora2019pixels,
  title={Pixels to perps: A deep learning algorithm for binary tactile seriation},
  author={Lepora, Nathan F and et al.},
  journal={IEEE Robotics and Automation Letters},
  volume={4},
  number={2},
  pages={2101--2107},
  year={2019}
}

@article{james2018slip,
  title={Slip detection with a biomimetic optical tactile sensor},
  author={James, Stephen and Lepora, Nathan F and et al.},
  journal={IEEE Robotics and Automation Letters},
  volume={3},
  number={3},
  pages={1506--1513},
  year={2018}
}

@inproceedings{calandra2018more,
  title={More than a feeling: Learning to grasp and regrasp using vision and touch},
  author={Calandra, Roberto and et al.},
  booktitle={2018 IEEE International Conference on Robotics and Automation (ICRA)},
  pages={3340--3347},
  year={2018}
}

@inproceedings{zhao2020lvis,
  title={LVIS: A large-scale video-based tactile dataset},
  author={Zhao, Siyuan and et al.},
  booktitle={Proceedings of the IEEE/CVF Conference on Computer Vision and Pattern Recognition},
  year={2020}
}

@inproceedings{ding2021sim,
  title={Sim-to-real transfer for robotic manipulation with tactile feedback},
  author={Ding, Zihan and et al.},
  booktitle={2021 IEEE/RSJ International Conference on Intelligent Robots and Systems (IROS)},
  year={2021}
}

\end{document}